\DocumentMetadata{%
	pdfstandard=A-3u,
	pdfversion=1.7,
	lang=en-US,
}

\documentclass[balance,colorlinks,upint,subscriptcorrection,varvw,mathalfa=cal=boondoxo,spanish,german,vietnamese,russian,greek]{asmeconf}

\usepackage{float} 
\usepackage{graphics}

\hypersetup{
pdfauthor={Bogdan Bogachov},
pdftitle={A Lightweight Multi-Expert Generative Language Model System for Engineering Information and Knowledge Extraction},
pdfkeywords={Large Language Model, Fine-tuning, Adaptation, Small Language Model, Small Language Graph, Generative AI},
pdfsubject = {Application and adaptation of LLMs and LLM systems in engineering},
}

\begin{document}



\title{A Lightweight Multi-Expert Generative Language Model System for Engineering Information and Knowledge Extraction}
 
%
%
%

\SetAuthors{Bogdan Bogachov\affil{1}, Yaoyao Fiona Zhao\affil{1}\CorrespondingAuthor{yaoyao.zhao@mcgill.ca}}

\SetAffiliation{1}{McGill University, Montreal, QC, Canada}


\maketitle



\keywords{Large Language Model, Fine-tuning, Adaptation, Small Language Model, Small Language Graph, Generative AI}


\begin{abstract}

Despite recent advancements in domain adaptation techniques for large language models, these methods remain computationally intensive, and the resulting models can still exhibit hallucination issues. Most existing adaptation methods do not prioritize reducing the computational resources required for fine-tuning and inference of language models. Hallucination issues have gradually decreased with each new model release. However, they remain prevalent in engineering contexts, where generating well-structured text with minimal errors and inconsistencies is critical. This work introduces a novel approach called the Small Language Graph (SLG), which is a lightweight adaptation solution designed to address the two key challenges outlined above. The system is structured in the form of a graph, where each node represents a lightweight expert—a small language model fine-tuned on specific and concise texts. The results of this study have shown that SLG was able to surpass conventional fine-tuning methods on the Exact Match metric by 3 times. Additionally, the fine-tuning process was 1.7 times faster compared to that of a larger stand-alone language model. These findings introduce a potential for small to medium-sized engineering companies to confidently use generative AI technologies, such as LLMs, without the necessity to invest in expensive computational resources. Also, the graph architecture and the small size of expert nodes offer a possible opportunity for distributed AI systems, thus potentially diverting the global need for expensive centralized compute clusters.

\end{abstract}


\section{Introduction}
\label{Introduction}

In recent years, Large Language Models (LLMs) have experienced a surge in popularity due to their ability to process and generate extensive amounts of data in response to user-defined queries. Major technology companies have been competing to deliver the most advanced LLMs on the market, resulting in models equipped with vast amounts of publicly available online knowledge. The most prominent examples of such systems in use are closed-source ChatGPT \cite{openai}, Gemini \cite{gemini}, and open-source Llama models \cite{touvron2302llama}. These systems can serve as effective assistants in domains grounded in well-established knowledge, where relevant information is readily or easily accessible through open-source data such as mathematics, law, and biology.

On the other hand, LLM systems may sometimes lack the necessary knowledge to answer a user query—particularly when the requested information was not included in the training data. To cope with this difficulty, agents were introduced. In general terms, agents \cite{sheng2024languagemodelspracticalselfimproving, xi2023risepotentiallargelanguage, Yao2023} act as "helpers" to LLM systems, capable of performing fact-checking, retrieving up-to-date and reliable information from the internet, and mitigating hallucination issues.

However, LLMs and LLM systems with agents struggle in narrow and specific domains such as design and manufacturing. As is widely known, the efficacy of LLMs is directly linked to the volume and quality of data available for training and fine-tuning. The key to producing efficient models is high-quality data \cite{luo2022data, han2020data}. Yet, taking various factors into account, including security \cite{behnia2022ew}, data in many design and manufacturing sub-fields is not publicly accessible, leading to challenges in obtaining domain-specific information. It can be argued that publicly available data is sufficient for developing state-of-the-art LLMs, and that transfer learning \cite{raffel2020exploring, radford2019language} enables near-optimal data processing and generation capabilities for end users. However, these models are not entirely reliable in specific applications and are prone to hallucinations even when agents are employed due to the inaccessibility of proprietary data.

Now, understanding the necessity of such systems in engineering domains is a critical aspect of this discussion. When properly adapted, these systems have the potential to substantially reduce the man-hours required for routine tasks (such as searching standard repair procedures for aerospace components), thereby freeing up the workforce to focus on more creative and value-added engineering activities. This could significantly increase the productivity of engineering firms. Furthermore, the financial aspect must also be considered. While readily available LLM models or systems could be adapted for engineering applications, most rely on costly cloud computing services or require the deployment of high-end on-premises servers. The majority of small to medium-sized engineering companies will not be able to afford such costly technologies. Therefore, there is a clear need for lightweight LLM adaptation techniques tailored to specific domains, aimed at reducing hallucinations and enhancing their accessibility for engineering applications.

In this research, the problem stated above is addressed by introducing SLG, a system comprised of transformer-based \cite{NIPS2017_3f5ee243} language model experts, which are based on fine-tuned Llama-3.2-1B-Instruct models \cite{Llama-3.2-1B-Instruct, grattafiori2024llama3herdmodels}. The reasoning behind choosing a graph system instead of fine-tuning a stand-alone LLM is due to the hallucination problem, since in any engineering domain, word inaccuracies or ambiguities are highly undesirable. Transformer-based \cite{NIPS2017_3f5ee243} models lack reasoning skills \cite{zevcevic2023causal} because, while being trained, they simply learn underlying word patterns in training data. Thus, during inference, the word generation process is purely probabilistic. The probabilistic nature of LLMs introduces a high risk of generating words that could not necessarily be related to the question of an engineer. One of the main reasons why this situation could happen is due to training data overlap. This issue, referred to as "knowledge overshadowing" \cite{zhang2024knowledgeovershadowingcausesamalgamated}, describes how overlapping contexts in the training data can blend together, making it difficult for an LLM to distinguish between identical or similar words with different meanings.

In SLG, the use of relatively small expert models, such as Llama-3.2-1B-Instruct \cite{Llama-3.2-1B-Instruct}, enables small to medium-sized engineering firms to deploy generative AI technologies locally. Additionally, the graph-based nature of SLG enhances text generation accuracy by leveraging expert nodes trained on focused, domain-specific data segments.

The remainder of the paper is structured as follows. Section \ref{Related work} discusses the related work. Section \ref{Methodology} explains in detail the proposed methodology and the architecture of the SLG system. Experiments are detailed in Section \ref{Experiments}. Limitations and future work are introduced in Section \ref{Limitations}. Finally, the conclusions and discussion are listed in Section \ref{Conclusions_discussion}.


\section{Related work}
\label{Related work}

This study proposes the following classification of technologies used to tackle the problem of LLM adaptation in engineering domains: prompt engineering, fine-tuning, and Retrieval-Augmented Generation (RAG).

\subsection{Prompt engineering}
\label{Prompt_engineering}

Prompt engineering offers several advantages, including easy access to preferred LLM systems, rapid interaction, swift generation of desired information, and the ability for users to focus on creative tasks rather than the meticulous process of searching for and extracting knowledge. Ready-to-use models are accessible online through platforms, such as OpenAI \cite{openai}, Gemini \cite{gemini}, etc. These platforms are user-friendly and provide access to their basic models free of charge. Studies conducted on prompt engineering \cite{Ma2023, Bouschery2023139, Korzynski202325} as a method to augment human knowledge have shown the usefulness of LLMs to tackle text generation tasks and speed up workflows. Among the advantages of prompt engineering are ease of access to the LLM systems of choice, fast interaction, quick generation of requested information, and the possibility for users to concentrate on creativity rather than on the scrutinized process of knowledge search and extraction. However, this method has significant drawbacks. LLM systems like ChatGPT \cite{openai} are prone to bias and hallucinations \cite{Bang}. Also, as specified in \cite{Ma2023, Bouschery2023139}, LLMs are sensitive to the quality of user prompts. Prompt sensitivity leads to high variability in LLM responses to similar questions that are phrased differently. Moreover, LLM systems lack the cognitive ability to truly understand context and rely solely on probability distributions when generating text \cite{zevcevic2023causal}. Agents \cite{sheng2024languagemodelspracticalselfimproving, xi2023risepotentiallargelanguage, Yao2023} offer a partial solution to the issues outlined above; however, they cannot address cases where user queries involve knowledge that is not accessible online.

\subsection{Fine-tuning}
\label{finet_tuning}

One approach to overcoming the limitation of inaccessible online knowledge is to ingest proprietary or non-public data into a pre-trained LLM. The most commonly known way of ingestion is fine-tuning. From a macro perspective, fine-tuning techniques can be classified into two major approaches: fine-tuning by means of modifying a base pre-trained model and fine-tuning by means of adding new layers or adapters on top of a base pre-trained model while keeping a base model unchanged.

The full fine-tuning method described in \cite{wei2021finetuned} proves its efficiency against prompt engineering. The authors used LaMDA-PT \cite{Thoppilan2022LaMDALM} as a backbone model. Its fine-tuned variant outperformed the backbone model by equipping it with additional knowledge. However, the study specifies several limitations. The most significant one is the high computational cost induced by updating all 137 billion parameters of the model.

In contrast, a notable example of LLM adaptation through the addition of extra layers atop a backbone model is Hierarchical Domain Adaptation (HDA), as introduced in \cite{chronopoulou2021efficient}. HDA \cite{chronopoulou2021efficient} leverages a pre-trained model and trains multiple domain-specific adapters, which are attached one at a time on top of the base model depending on a task being performed. Another similar method is Low-Rank Adaptation of LLMs, or LoRA \cite{Hu2022}. Similar to HDA \cite{chronopoulou2021efficient}, LoRA \cite{Hu2022} introduces additional layers on top of a frozen backbone model. LoRA employs a bottleneck architecture that substantially reduces the number of trainable parameters, enabling faster training and inference with minimal added latency.

It is worth noting that the above-mentioned reported literature lacks hallucination tests. Since all of the described methods involve fine-tuning LLMs on data from whole domains, knowledge overshadowing \cite{zhang2024knowledgeovershadowingcausesamalgamated} mentioned in Section \ref{Introduction} could occur, thus invoking hallucinations.

\subsection{RAG systems}
\label{RAG}

One of the most impactful technologies potentially able to solve the hallucination phenomenon in LLMs is RAG. Originally introduced in \cite{Lewis2020}, this method was welcomed by researchers and professionals around the world not only as a way to fight hallucinations but also as a strong option to augment knowledge of any LLM \cite{Li20256705, improving_llms_applications_in_biomed}. 

To achieve this, RAG chunks textual information, converts chunks to dense vectors, and stores them in a vector database. During inference, relevant chunks in the form of vectors are retrieved. Retrieval is achieved by comparing a vectorized user query with vectors in the vector database created previously. Top-k vectors are then selected to be passed as context to a generator LLM, which composes a response. This approach can significantly enhance the knowledge of an LLM and reduce hallucinations by enabling access to a dynamically updated vector database containing the most current information.

One of the latest developments in RAG was shared in \cite{Raina2024219}. This research introduces a retrieval method based on questions using atomic units to improve the retrieval step in RAG systems. This approach enhances recall by breaking text chunks into smaller atomic statements and generating synthetic questions to match user queries more accurately. However, an assumption is made that each query has a single answerable chunk. Also, it does not handle multi-hop retrieval and has only been tested on small-scale datasets.

Consequently, it implies that RAG is not a panacea for all deficiencies of LLMs. This methodology struggles with noisy data and is sporadically incapable of providing negative rejection, an ability to refuse answering a question when retrieved documents lack relevant information \cite{Chen202417754}.

\subsection{LLMs in engineering}
\label{llms_in_engineering}

One of the most recent works devoted to adapting LLMs in engineering domains employing prompt engineering \cite{Chen2024} introduces a novel method to extract aviation accident causality information. The approach presented in this paper is compared with existing LLM-based information extraction methods and is reported to outperform them by achieving higher accuracy, requiring less annotated data, and handling unstructured text more effectively. However, this method struggles with processing ambiguous texts and requires high computational resources.

LLMs' fine-tuning, presented by \cite{Kumar20251873}, showcases a solid method tailored to solve engineering problems. This paper introduces a set of MechBERT models, LLMs based on Bidirectional Encoder Representations for Transformer (BERT). The models were pre-trained on stress–strain scientific literature and further fine-tuned for general English-language question-answering tasks to improve information extraction of mechanical properties. The resultant models outperformed other models in the BERT family while being smaller and faster. However, despite the performance increase in the domain of interest, the models showed limited performance on general-language tasks.

Finally, \cite{Joshi2024} offers promising insights into using RAG in engineering. This paper proposes a RAG-based tool to extract information from documents encompassing multiple domains. The tool provides a high level of semantic understanding, flexibility in domain adaptation, and integration. Nevertheless, the proposed technique is overly reliant on complex models, and it lacks standardized evaluation metrics.

Motivated by the limitations outlined in the preceding subsections, there is a clear need to develop a method that combines computational efficiency with high accuracy while effectively addressing domain-specific tasks.

\section{Methodology}
\label{Methodology}

The methodology used to create SLG is split into two main portions: dataset preparation and the SLG system construction.

\subsection{Dataset}
\label{dataset}

Since this work is aimed at finding a lightweight LLM adaptation solution tailored to maximize accuracy while generating engineering data, any text-based engineering document is sufficient as a dataset. In this research, a Structural Repair Manual (SRM) of Cessna aircraft is used \cite{cesna_srm}.

\begin{figure}[h]
    \centering
    \includegraphics[width=0.45
    \textwidth]{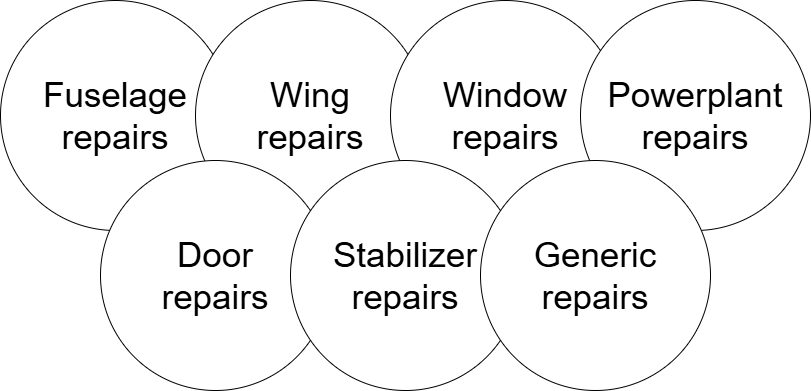}
    \caption{Data overlapping illustration.}
    \label{dataovershadowing}
\end{figure}

Increasing LLM generation accuracy could involve multiple approaches. One of them is aiming to reduce hallucinations. As it was mentioned earlier, one of the reasons for the hallucination phenomenon is data overshadowing \cite{zhang2024knowledgeovershadowingcausesamalgamated}. In an oversimplified way, this phenomenon can be described as data overlapping, as shown in Figure \ref{dataovershadowing}.

An example of such overlapping could happen when two or more engineering procedures have identical beginnings but different endings, as shown in Table \ref{sentence_overlap}.

\begin{table}[H]
    \caption[Table]{Example of data overlapping from \cite{cesna_srm}.}
    \label{sentence_overlap}
    \centering{
        \begin{tabular}{{p{2cm} p{6cm}}}
            \toprule
             & \multicolumn{1}{c}{Sentence} \\ 
            \midrule
            Sentence 1 & \textbf{Damage which would involve a} typical skin repair can be described as damage that requires modification, such as material replacement or patching. \\ 
            Sentence 2 & \textbf{Damage which would involve a} control surface repair: After the repair is completed, the control surface
    balance must be checked as described in Flight Control Surface Balancing. \\ 
            \bottomrule
        \end{tabular}
    }
\end{table}

\begin{figure}
    \centering
    \includegraphics[width=0.45
    \textwidth]{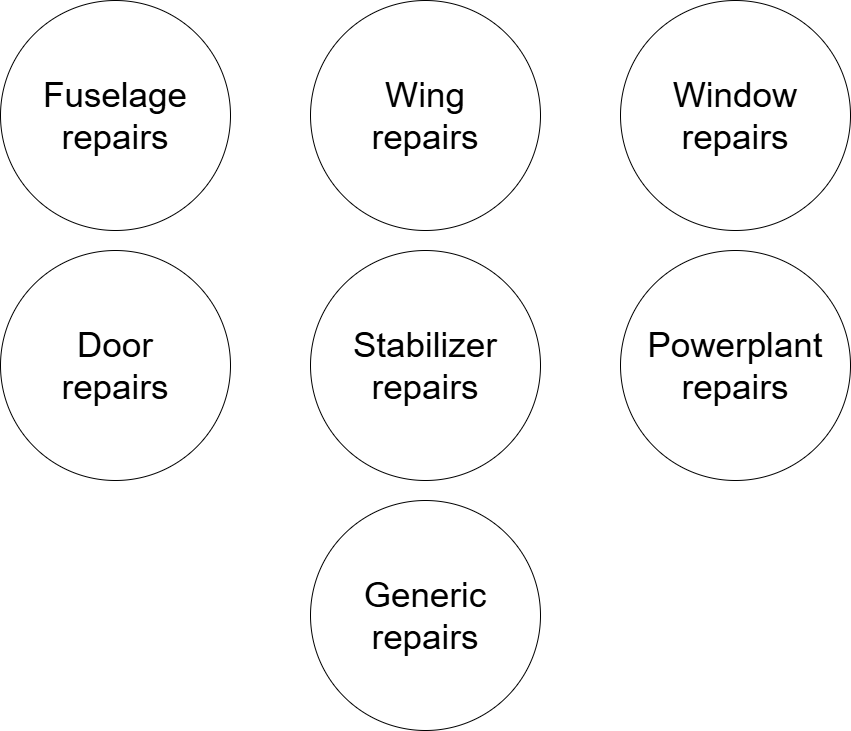}
    \caption{Schematic representation of isolated training data.}
    \label{isolated}
\end{figure}

To avoid overlapping, training data chunks were isolated from each other. A schematic example of an ideal training dataset split would look as shown in Figure \ref{isolated}, where each bubble represents a small chunk of the whole training dataset. Each chunk is used to fine-tune only one expert in the SLG system. This way, each expert receives isolated knowledge, thus eliminating data overlap.

To achieve such data division, each training data chunk has to have a logical beginning and a logical ending. Usually, text-based engineering documentation has a well-defined structure split by sections and subsections; Cessna's SRM \cite{cesna_srm} is not an exception. This feature of engineering documentation simplifies the data preparation process. The text is split into chunks by subsections. Subsequently, each chunk is fed into Llama-3.3-70B-Instruct LLM \cite{Llama-3.3-70B-Instruct}, asking it to generate questions for the text. Thus, question-answer pairs are created, which are used for model fine-tuning and testing.

It is important to note that this data chunking method is well-suited to most engineering documentation due to its structured nature, making SLG applicable across a wide range of engineering domains.

Also, it is essential to highlight that all image data was removed from the dataset due to the text-only focus of this specific research.

\subsection{SLG}
\label{slg}

\begin{figure*}[hbtp]
    \centering
    \includegraphics[width=0.9
    \textwidth]{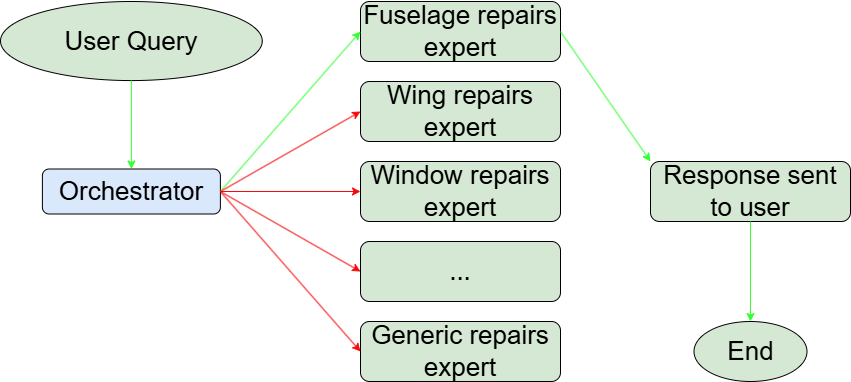}
    \caption{Small Language Graph.}
    \label{slg_figure}
\end{figure*}

The methodology used to build the SLG system is based on graphs, as shown in Figure \ref{slg_figure}. In the flowchart, it is assumed that the user's query is about fuselage repairs. The process follows the green arrows. The query is first directed to the orchestrator, which then queries the fuselage repairs expert. A response is returned to the user, and the process concludes at the end block.

The system is built on the Llama-3.2-1B-Instruct \cite{Llama-3.2-1B-Instruct} as its backbone LLM, which is fine-tuned using LoRA \cite{Hu2022} to serve both as the orchestrator and the expert nodes.

A dataset used for the orchestrator differs from the one used for experts. In both cases, the datasets share identical questions; however, the answers used for expert fine-tuning are actual engineering procedures, while for the orchestrator, the answers are the expert names. The expert names bear the names of the engineering document subsections. This approach allows the orchestrator to directly return the name of an appropriate expert and send the user's query to it. Refer to a question-answer example in Table \ref{qa_example}.

\setlength{\tabcolsep}{6pt} 
\begin{table*}[h]
    \caption{SLG experts and orchestrator question-answer pairs example \cite{cesna_srm}.}
    \label{qa_example}
    \centering{%
        \renewcommand{\arraystretch}{1.2} 
        \begin{tabular*}{0.9\textwidth}{@{\hspace*{1.5em}}@{\extracolsep{\fill}}m{2.9cm}m{5.2cm}m{6.2cm}@{\hspace*{1.5em}}}
            \toprule
            \multicolumn{1}{@{\hspace*{1.5em}}c}{\textbf{Orchestrator answer}} &
            \multicolumn{1}{c}{\textbf{Common question}} &
            \multicolumn{1}{c@{\hspace*{1.5em}}}{\textbf{Expert answer}} \\ 
            \midrule
            \centering WING DAMAGE\newline CLASSIFICATION &
            What are the key factors that determine whether damage to the wing fuel bay spars or ribs can be addressed through repair or requires replacement, considering the criteria outlined for negligible, repairable, and replacement-necessitating damage? &
            Wing Fuel Bay Spars/Rib Damage Criteria. Negligible damage: Any smooth dents in the wing fuel spar and ribs that have no evidence of tears, cracks, or penetrations – which are not stress wrinkles and do not change (oil can, or pop in and out) with internal pressure – are considered negligible damage... \\ 
            \bottomrule
        \end{tabular*}
    }
\end{table*}

To fine-tune experts, the same backbone Llama-3.2-1B-Instruct \cite{Llama-3.2-1B-Instruct} LLM is used. The model is fine-tuned separately on each isolated dataset described in Subsection \ref{dataset} using LoRA \cite{Hu2022}. The fine-tuned models are then connected using a graph approach utilizing the LangGraph library \cite{langgraph}; thus, each model is represented by a node, and the orchestrator extends edges to each of the experts.

To perform inference within the SLG system, the orchestrator receives a user query, processes it, and routes it to the most relevant expert node for response generation. A chosen expert produces an answer, which is returned to the user.

A detailed evaluation of the proposed method is presented in Section \ref{Experiments}, with Table \ref{hyperparams} providing a comprehensive list of all hyperparameters used.

\section{Experiments}
\label{Experiments}

This section describes the experimentation setup, followed by the fine-tuning strategy of all tested models.

\subsection{Experimentation setup}
\label{experimentation_setup}

It is implied by model metrics on different benchmarks that Llama-3.1-8B-Instruct LLM \cite{Llama-3.1-8B-Instruct} exhibits better performance than Llama-3.2-1B-Instruct LLM \cite{Llama-3.2-1B-Instruct}.

Since SLG is based on a small LLM - Llama-3.2-1B-Instruct \cite{Llama-3.2-1B-Instruct}, to prove the potential of SLG, it is proposed to compare it with fine-tuned Llama-3.1-8B-Instruct LLM \cite{Llama-3.1-8B-Instruct} and fine-tuned Llama-3.2-1B-Instruct LLM \cite{Llama-3.2-1B-Instruct}. The core objective of this experimental setup is to demonstrate that the fine-tuned multi-expert SLG system outperforms both a larger stand-alone fine-tuned Llama-3.1-8B-Instruct LLM \cite{Llama-3.1-8B-Instruct} and a size-matched stand-alone fine-tuned Llama-3.2-1B-Instruct LLM \cite{Llama-3.2-1B-Instruct}.

All models are tested using a test dataset described in Subsection \ref{dataset} by comparing generated answers to ground truth answers.

ROUGE-L, Exact Match (EM), and METEOR are used as evaluation metrics in this research, where ROUGE-L measures the longest common subsequence between the generated and reference texts, EM checks for an exact string match between the prediction and the reference, and METEOR evaluates based on unigram matches while considering synonyms, stemming, and word order.

\subsection{Fine-tuning strategy}
\label{finetuning_strategy}

LoRA \cite{Hu2022} is chosen as a fine-tuning technique in this research. The finetuning pipeline and hyperparameters are shared among all models, namely, Llama-3.2-1B-Instruct LLM \cite{Llama-3.2-1B-Instruct}, Llama-3.1-8B-Instruct LLM \cite{Llama-3.1-8B-Instruct}, SLG. This approach allows a fair comparison by fixing all variables.

The experiments in this study are divided into four categories, each focusing on tuning a specific hyperparameter in the following sequence: learning rate, LoRA \cite{Hu2022} rank, gradient accumulation, and LoRA \cite{Hu2022} alpha.

Table \ref{hyperparams_tuned} lists all combinations of tuned hyperparameters. The values in bold indicate which hyperparameter is tuned at each specific row. After tuning, the values exhibiting the best performance are fixed and highlighted in green. All other hyperparameters are fixed and listed in Table \ref{hyperparams}.

\begin{table*}
    \captionsetup{width=0.9\textwidth, justification=raggedright}
    \caption{Tuned hyperparameters.}
    \centering
    \begin{tabular}{p{3cm} p{3cm} p{2.5cm} p{4cm} p{2.5cm}}
        \toprule
        
        \textbf{Experiment \#}
        & \textbf{Learning rate}
        & \textbf{LoRA rank}
        & \textbf{Gradient accumulation}
        & \textbf{LoRA alpha} \\
        \midrule
        1 & \textbf{1e-5}           & 4                    & 2                               & 8 \\
        2 & \textbf{1e-4}           & 4                    & 2                               & 8 \\
        3 & {\textcolor{ForestGreen}{\textbf{1e-3}}}           & 4                    & 2                               & 8 \\
        4 & {\textcolor{ForestGreen}{1e-3}}  & \textbf{8}           & 2                               & 8 \\
        5 & {\textcolor{ForestGreen}{1e-3}}  & {\textcolor{ForestGreen}{\textbf{16}}} & 2                               & 8 \\
        6 & {\textcolor{ForestGreen}{1e-3}}  & \textbf{32}          & 2                               & 8 \\
        7 & {\textcolor{ForestGreen}{1e-3}}  & {\textcolor{ForestGreen}{16}} & {\textcolor{ForestGreen}{\textbf{2}}} & 8 \\
        8 & {\textcolor{ForestGreen}{1e-3}}  & {\textcolor{ForestGreen}{16}} & \textbf{4}                      & 8 \\
        9 & {\textcolor{ForestGreen}{1e-3}}  & {\textcolor{ForestGreen}{16}} & \textbf{8}                      & 8 \\
        10 & {\textcolor{ForestGreen}{1e-3}} & {\textcolor{ForestGreen}{16}} & {\textcolor{ForestGreen}{2}} & {\textcolor{ForestGreen}{\textbf{8}}} \\
        11 & {\textcolor{ForestGreen}{1e-3}} & {\textcolor{ForestGreen}{16}} & {\textcolor{ForestGreen}{2}} & \textbf{16} \\
        12 & {\textcolor{ForestGreen}{1e-3}} & {\textcolor{ForestGreen}{16}} & {\textcolor{ForestGreen}{2}} & \textbf{32} \\
        13 & {\textcolor{ForestGreen}{1e-3}} & {\textcolor{ForestGreen}{16}} & {\textcolor{ForestGreen}{2}} & \textbf{64} \\
        \bottomrule
    \end{tabular}
    \label{hyperparams_tuned}
\end{table*}

\begin{table}[h]
    \caption{Hyperparameters used for fine-tuning.}
    \centering
    \begin{tabular}{p{5cm} p{3cm}}
        \toprule
        \textbf{Hyperparameter} & \textbf{Value} \\
        \midrule
        LoRA alpha & Refer to Table \ref{hyperparams_tuned} \\
        LoRA r & Refer to Table \ref{hyperparams_tuned} \\
        LoRA dropout & 0.05 \\
        LoRA task_type & CAUSAL_LM \\

        learning_rate & Refer to Table \ref{hyperparams_tuned} \\
        gradient_accumulation_steps & Refer to Table \ref{hyperparams_tuned} \\
        weight_decay & 0.001 \\
        label_smoothing_factor & 0.01 \\
        optim & adamw_torch \\
        num_train_epochs & 25 (early stopped) \\
        early_stopping_patience & 3 \\
        eval_strategy & epoch \\
        save_strategy & epoch \\
        fp16 & True \\
        per_device_train_batch_size & 2 \\
        per_device_eval_batch_size & 2 \\
        adam_beta1 & 0.9 \\
        adam_beta2 & 0.999 \\
        max_grad_norm & 0.5 \\
        warmup_ratio & 0.03 \\
        lr_scheduler_type & linear \\
        load_best_model_at_end & True \\
        save_total_limit & 4 \\
        \bottomrule
    \end{tabular}
    \label{hyperparams}
\end{table}

For the full fine-tuning pipeline refer to ‘finetune.py’ \cite{finetune_slg} module in the SLG repository.

\subsection{Results}
\label{results}

Overall, the initial experimental results demonstrate the efficiency of the SLG system, built on smaller Llama-3.2-1B-Instruct models \cite{Llama-3.2-1B-Instruct}, outperforming both the stand-alone Llama-3.1-8B-Instruct \cite{Llama-3.1-8B-Instruct} and the stand-alone Llama-3.2-1B-Instruct \cite{Llama-3.2-1B-Instruct} models.

\begin{figure*}
    \centering
    \includegraphics[width=0.85\textwidth]{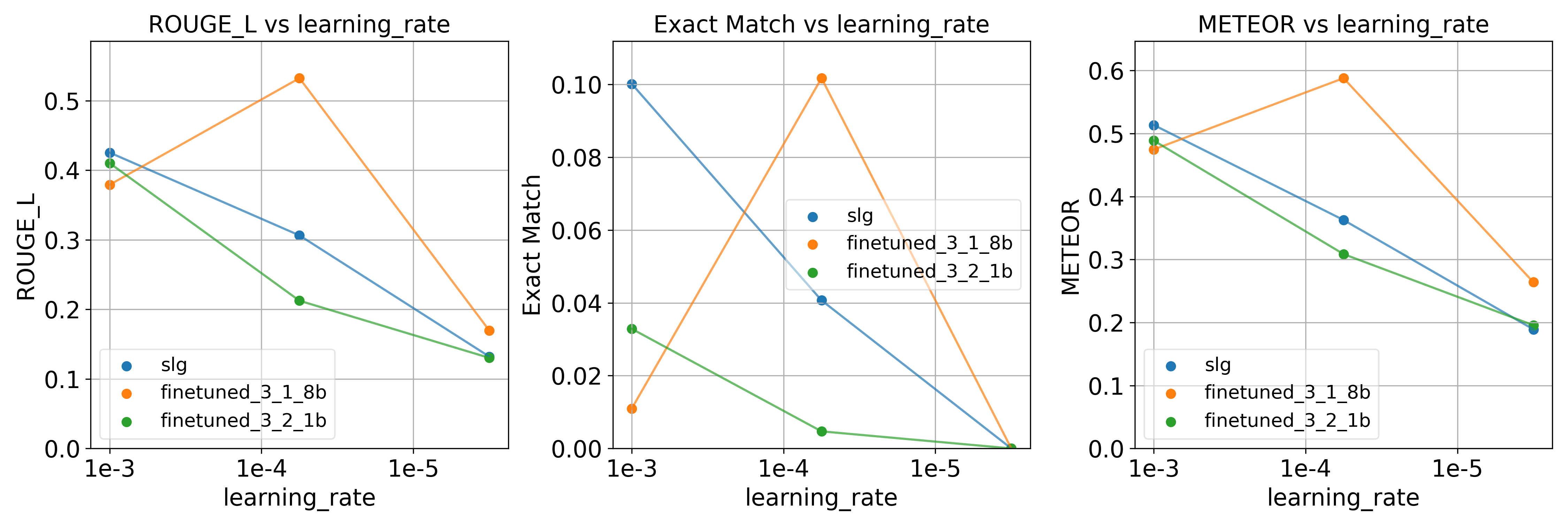} \\
    \includegraphics[width=0.85\textwidth]{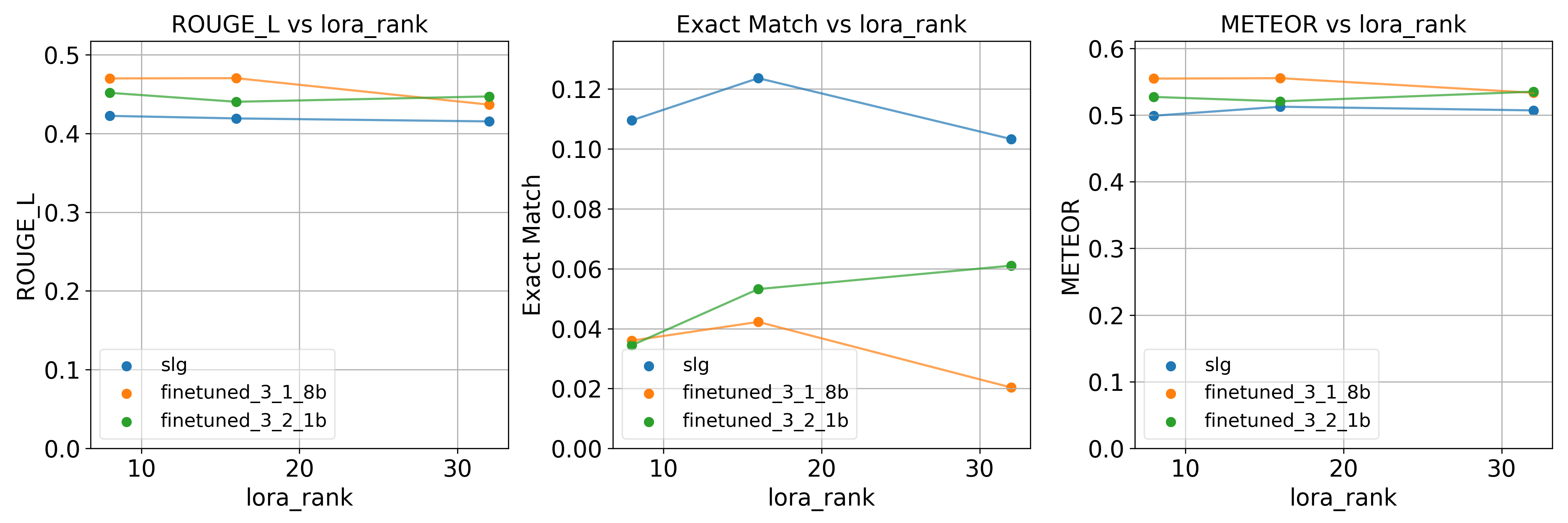} \\
    \includegraphics[width=0.85\textwidth]{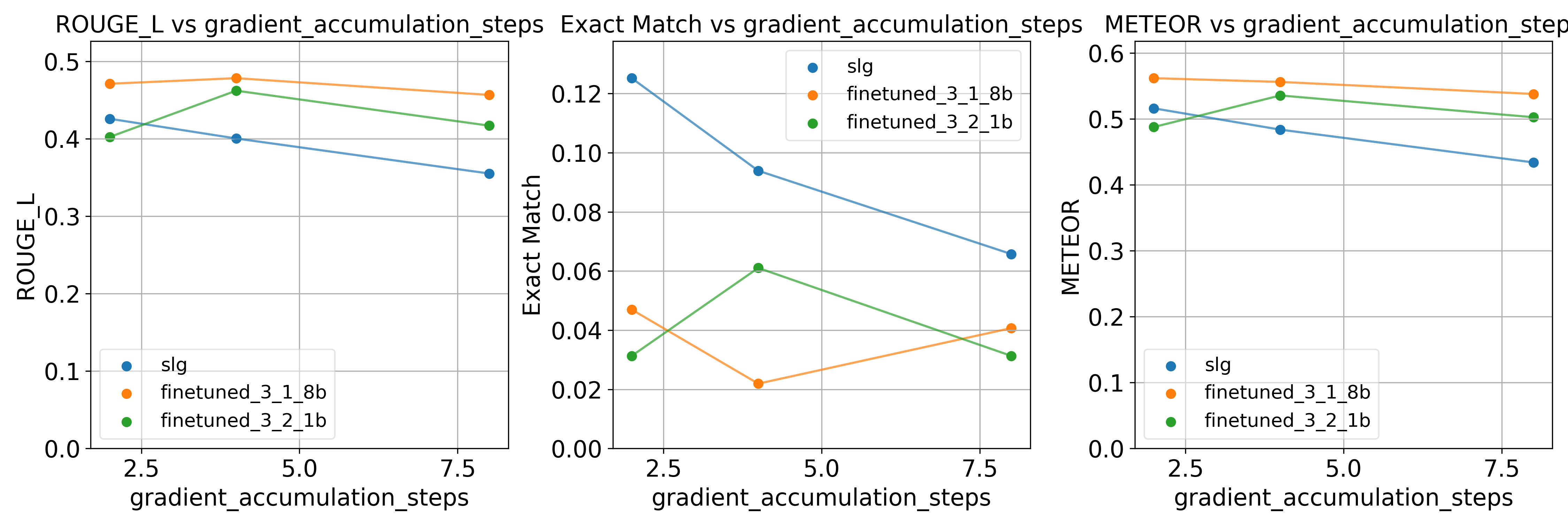} \\
    \includegraphics[width=0.85\textwidth]{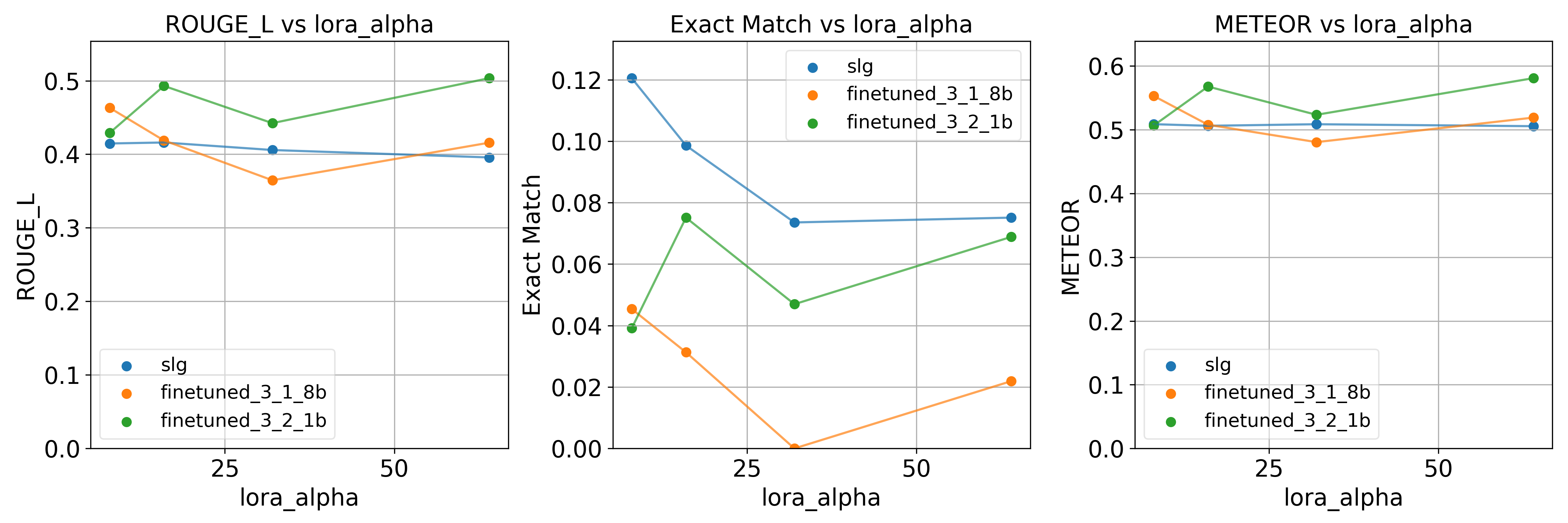}
    \caption{Experiment Charts.}
    \label{charts}
\end{figure*}

Figure \ref{charts} illustrates the experimentation evolution. The charts are organized as follows: rows iterate over tuned hyperparameters, while columns iterate over evaluation metrics. Rows one to four showcase comparisons of learning rate, LoRA rank, gradient accumulation, and LoRA alpha against the corresponding metrics. Columns one to three depict comparisons of ROUGE-L, EM, and METEOR across the corresponding hyperparameters.

Table \ref{results_table} showcases the best experiment results, where R-L, EM, and M stand for ROUGE-L, EM, and METEOR, respectively. While ROUGE-L and METEOR metrics demonstrate similar performance on all compared models, the EM metric indicates that SLG can achieve 3 times better results. Among the three used metrics, EM is the most powerful indication that SLG has the potential to better resist hallucinations by producing text exactly matching the engineering ground truth answers.

In addition, all SLG experts and its orchestrator are trained 1.7 times faster than one stand-alone Llama-3.1-8B-Instruct LLM \cite{Llama-3.1-8B-Instruct}, as demonstrated in Table \ref{finetuning_time}.

\begin{table}[h]
    \centering
    \caption{Fine-tuning time comparison.}
    \begin{tabular}{p{2.8cm} p{4.5cm}}
        \toprule
        \textbf{Model} & \multicolumn{1}{c}{\textbf{Average fine-tuning time}} \\
        \midrule
        SLG & \multicolumn{1}{c}{3475 seconds} \\
        Llama-3.1-8B \cite{Llama-3.1-8B-Instruct} & \multicolumn{1}{c}{5891 seconds} \\
        Llama-3.2-1B \cite{Llama-3.2-1B-Instruct} & \multicolumn{1}{c}{2163 seconds} \\
        \bottomrule
    \end{tabular}
    \label{finetuning_time}
\end{table}

\begin{table}[h]
    \centering
    \caption{Best experiment metrics.}
    \begin{tabular}{p{2.8cm} p{1.5cm} p{1.5cm} p{1.5cm}}
        \toprule
        \textbf{Model} & \textbf{R-L} & \textbf{EM} & \textbf{M} \\
        \midrule
        SLG & 0.41 & 0.12 & 0.50 \\
        Llama-3.1-8B \cite{Llama-3.1-8B-Instruct} & 0.46 & 0.05 & 0.55 \\
        Llama-3.2-1B \cite{Llama-3.2-1B-Instruct} & 0.43 & 0.04 & 0.51 \\
        \bottomrule
    \end{tabular}
    \label{results_table}
\end{table}

Furthermore, SLG has the potential to exhibit better performance on all three metrics if the orchestrator node is improved. It was discovered that the orchestrator did not always direct user queries to the appropriate expert, thus decreasing the performance of SLG. The success rate of the orchestrator is approximately 70\% and is subject to improvement in the future iterations.

Lastly, yet importantly, SLG is able to be fine-tuned and inferred on only one NVIDIA RTX 4090 (24GB VRAM) Graphics Processing Unit (GPU), which makes the system undoubtedly lightweight.

\section{Limitations and future work}
\label{Limitations}

Although SLG demonstrated significant potential in generating engineering texts, it has certain limitations and requires future adjustments.

One notable constraint of this research is its limit to only two models for comparisons, namely, Llama-3.1-8B-Instruct LLM \cite{Llama-3.1-8B-Instruct} and Llama-3.2-1B-Instruct LLM \cite{Llama-3.2-1B-Instruct}. It is planned to conduct more extensive comparisons by including the bigger Llama-3.3-70B-Instruct LLM \cite{Llama-3.3-70B-Instruct} and RAG \cite{Lewis2020}. As described in Subsection \ref{RAG}, RAG is a very powerful technique that enables LLMs to access up-to-date information and augment their contexts before generating text. Llama-3.3-70B-Instruct LLM \cite{Llama-3.3-70B-Instruct}, on the other hand, demonstrates better results than GPT-4o on most benchmarks \cite{llama_website_with_metrics}; thus, it is a great candidate for comparisons. Also, the experimentation in this research focuses on tuning 4 hyperparameters only, while it is beneficial to extend the experimentation towards other potentially significant hyperparameters, namely, weight decay, learning rate scheduler, warmup ratio, and max gradient norm.

Another shortcoming lies in the limited hallucinations check. This study uses EM as a prevailing metric to showcase the superiority of SLG in resisting hallucinations in comparison to stand-alone LLMs; however, human evaluation and fact-checking could be a more exhaustive way to estimate how well SLG can avoid hallucinations.

A further limitation involves the absence of images in the training data due to the pure text-based focus of the study. It is an important aspect to consider in future works, since image data is essential in engineering.

It is important to acknowledge that the proposed version of SLG is not a full-scale chatbot, does not have memory, and does not keep conversational context. Each user query is a stand-alone question that does not lead to further communication after receiving an answer from the system. Also, as it was mentioned in Subsection \ref{results}, the orchestrator node does not always direct user queries to an appropriate expert. This issue could be overcome by converting SLG into a full-scale chatbot system, which would equip a user with the possibility to send clarifying prompts to the system and provide the orchestrator with the necessary information to make a proper decision. Also, an aggregator node could be added to the pipeline to collect text generated by experts into one piece of information in cases when the orchestrator would split a user query among multiple experts. A generic expert node could be a solid addition to the system too, in cases when the orchestrator would not find an appropriate expert at all.

\section{Conclusions and discussion}
\label{Conclusions_discussion}

This research proposes a lightweight SLG system tailored for engineering domains to enhance engineers’ knowledge and accelerate their workflows. By offloading repetitive tasks, the system enables engineers to focus on more creative and value-driven activities.

SLG employs ultra-small language models as nodes within a graph-based architecture. This design has demonstrated both efficiency and strong potential for mitigating hallucinations in LLMs by constraining each expert node to a narrowly defined knowledge domain. This knowledge isolation strategy minimizes data overlap, thereby reducing the risk of hallucinations. Using EM as the primary evaluation metric, SLG achieved results three times better than those of the larger stand-alone model, Llama-3.1-8B-Instruct.

As it was reported in \ref{experimentation_setup} that the Llama-3.1-8B-Instruct LLM outperforms the Llama-3.2-1B-Instruct LLM when used individually. Therefore, the threefold performance improvement achieved by SLG is particularly significant—it demonstrates that a system composed of multiple smaller and individually less capable Llama-3.2-1B-Instruct models can collectively outperform a much larger standalone model. Moreover, despite comprising multiple expert models, SLG achieves 1.7 times faster training than the Llama-3.1-8B-Instruct and requires substantially fewer computational resources, owing to the lightweight nature of its constituent models.

This finding opens the door to building larger, more scalable systems based on the SLG architecture. In particular, it points to the potential of distributed AI systems composed of small language models, such as Llama-3.2-1B-Instruct, where individual users contribute expert nodes running on personal devices like laptops or smartphones. Given that these expert models require minimal computational resources, the network can scale virtually without limit. Such an approach could eventually reduce the reliance on expensive compute clusters, shifting the paradigm toward decentralized AI infrastructure. This vision draws parallels with existing distributed systems, such as peer-to-peer file sharing enabled by the BitTorrent protocol \cite{torrent}.

\section{Acknowledgment}
\label{Acknowledgment}

This work is funded by McGill Engineering Doctoral Award (MEDA) with additional funding support from Natural Sciences and Engineering Research Council of Canada Discovery Grant RGPIN-2018-05971.

The authors thank Digital Research Alliance of Canada for providing computational resources.

\bibliographystyle{asmeconf}  

\end{document}